\documentclass{article}
\usepackage{ijcai23}

\usepackage{times}
\usepackage{soul}
\usepackage{url}
\usepackage[hidelinks]{hyperref}
\usepackage[utf8]{inputenc}
\usepackage[small]{caption}
\usepackage{graphicx}
\usepackage{amsmath}
\usepackage{amsthm}
\usepackage{booktabs}
\usepackage{algorithm}
\usepackage{algorithmic}
\usepackage[switch]{lineno}

\usepackage{bm}
\usepackage{amssymb}
\usepackage[table]{xcolor}
\usepackage{pgfplots}
\usepackage{subcaption}
\usepackage{tikz}
\usepackage{multirow}

\pgfplotsset{compat=newest}

\definecolor{mygray}{gray}{.9}
\definecolor{sota_blue}{HTML}{0071bc}
\makeatletter
\newcommand{\thickhline}{%
	\noalign {\ifnum 0=`}\fi \hrule height 1pt
	\futurelet \reserved@a \@xhline
}
\makeatother
\newcommand{\tablestyle}[2]{\setlength{\tabcolsep}{#1}\renewcommand{\arraystretch}{#2}\centering\footnotesize}

\title{Enriching Phrases with Coupled Pixel and Object Contexts for\\ Panoptic Narrative Grounding}

\author{
Tianrui Hui$^{1,2,5}$
\and
Zihan Ding$^{3,5}$\and
Junshi Huang$^5$\thanks{Corresponding author}\and
Xiaoming Wei$^5$\and
Xiaolin Wei$^5$\and\\
Jiao Dai$^{1,2}$\and
Jizhong Han$^{1,2}$\And
Si Liu$^{3,4}$
\affiliations
$^1$Institute of Information Engineering, Chinese Academy of Sciences\\
$^2$School of Cyber Security, University of Chinese Academy of Sciences\\
$^3$Institute of Artificial Intelligence, Beihang University\\
$^4$Hangzhou Innovation Institute, Beihang University\\
$^5$Meituan
}

\begin{document}

\maketitle

\begin{abstract}
    Panoptic narrative grounding (PNG) aims to segment things and stuff objects in an image described by noun phrases of a narrative caption.
    As a multimodal task, an essential aspect of PNG is the visual-linguistic interaction between image and caption.
    The previous two-stage method aggregates visual contexts from offline-generated mask proposals to phrase features, which tend to be noisy and fragmentary.
    The recent one-stage method aggregates only pixel contexts from image features to phrase features, which may incur semantic misalignment due to lacking object priors.
    To realize more comprehensive visual-linguistic interaction, we propose to enrich phrases with coupled pixel and object contexts by designing a Phrase-Pixel-Object Transformer Decoder (PPO-TD), where both fine-grained part details and coarse-grained entity clues are aggregated to phrase features.
    In addition, we also propose a Phrase-Object Contrastive Loss (POCL) to pull closer the matched phrase-object pairs and push away unmatched ones for aggregating more precise object contexts from more phrase-relevant object tokens.
    Extensive experiments on the PNG benchmark show our method achieves new state-of-the-art performance with large margins.
\end{abstract}

\begin{figure}[!t]
  \begin{center}
     \includegraphics[width=\linewidth]{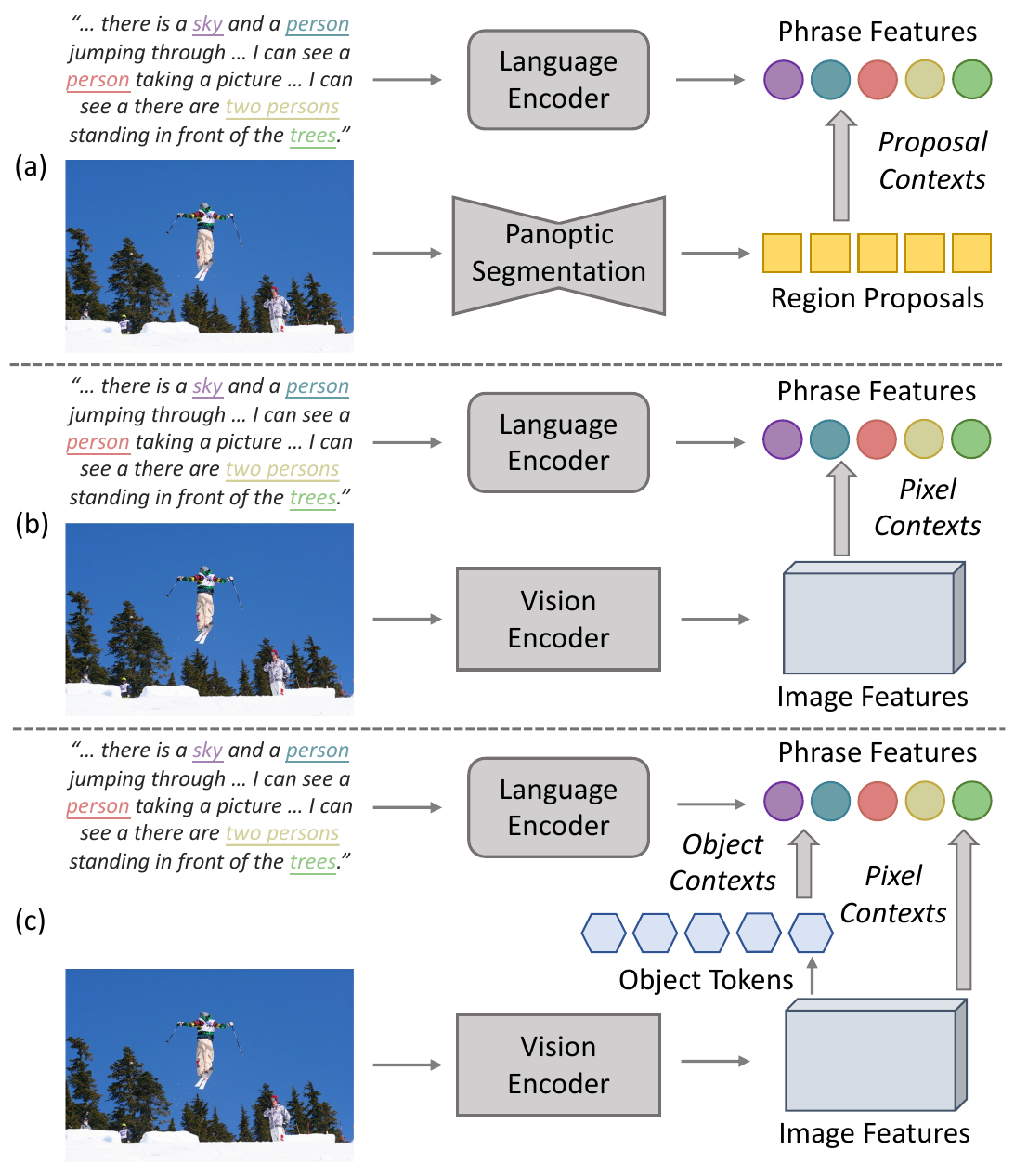}
  \end{center}
     \caption{Comparison of visual-linguistic interaction schemes between previous methods and ours. (a) The previous two-stage method aggregates proposal contexts from offline-generated mask region proposals, which tend to be fragmentary and noisy. (b) The previous one-stage method aggregates pixel contexts directly from image features, but lacking object-level contexts inclines to incur semantic misalignment between phrases and pixels. (c) Our method proposes to enrich phrases with coupled pixel and object contexts containing both fine-grained part details and coarse-grained entity clues, forming more comprehensive visual-linguistic interaction.}
  \label{fig:intro}
\end{figure}

\section{Introduction}
\label{sec:intro}

Panoptic narrative grounding (PNG)~\cite{gonzalez2021panoptic} is an emerging task whose goal is to predict the panoptic segmentation of things and stuff objects described by noun phrases in a narrative caption.
Compared with relevant tasks like referring image segmentation~\cite{yang2022lavt} and phrase grounding~\cite{mu2021disentangled}, PNG not only extends the grounding granularity from sentences to phrases but also yields finer results of segmentation masks rather than bounding boxes, facilitating a more comprehensive visual-linguistic alignment.
Benefiting from these properties, PNG enjoys a wide range of potential applications such as image editing, intelligent robots, and human-computer interaction.

Considering the multimodal nature of the PNG task, one of the essential components of PNG methods is the visual-linguistic interaction between image and caption.
As shown in Figure~\ref{fig:intro}~(a), the first work~\cite{gonzalez2021panoptic}, which proposes the PNG task, also designs a two-stage baseline method.
It first leverages an off-the-shelf panoptic segmentation model~\cite{kirillov2019pfpn} to generate mask region proposals for the input image, then performs matching between features of phrases and proposals.
Herein, the visual-linguistic interaction is realized by cross-attention~\cite{vaswani2017attention} between phrases and region proposals, where proposal contexts are aggregated to the phrase features for better matching in the second stage.
However, the region proposals are usually inaccurate segments, so the aggregated proposal contexts may provide fragmentary and noisy visual object information to the phrase features.
Recently, PPMN~\cite{ding2022ppmn} proposes a one-stage method that lets each phrase directly find its matched pixels to avoid the offline proposal generation process.
For visual-linguistic interaction, the one-stage method in Figure~\ref{fig:intro}~(b) aggregate pixel-level visual contexts directly from the grid-form image features to the phrase features.
Though pixel contexts contain fine-grained visual details, the absence of object-level visual contexts inclines to incur semantic misalignment between abstract noun phrases and concrete image pixels, yielding sub-optimal object delineation.

Given the above discussion, the visual-linguistic interaction schemes of previous methods still limit the discriminative ability of phrase features due to insufficient visual contexts, albeit their strong performances.
Therefore, in this paper, we propose to enrich phrases with coupled pixel and object contexts for more comprehensive visual-linguistic interaction as illustrated in Figure~\ref{fig:intro} (c).
On the one hand, pixel contexts aggregated directly from image features contain more fine-grained part details, which can assist phrases in predicting finer segmentation masks.
On the other hand, object contexts, which are also derived from image features, focus more on coarse-grained entity perception and can facilitate the distinguishment of corresponding referred objects for different phrases.
By coupling the pixel and object contexts, the discriminative ability of phrases could be further enhanced to excavate and match with more relevant visual objects.
Concretely, our method first extracts image and phrase features by vision and language encoders.
Then, phrases and a set of learnable object tokens are concatenated and fed into our proposed Phrase-Pixel-Object Transformer Decoder (PPO-TD) as queries, in which multi-scale image features from a pixel decoder serve as keys and values.
Through consecutive masked cross-attention layers, phrases aggregate pixel contexts from image features to obtain fine-grained part details.
Meanwhile, the representations of object tokens are progressively learned to perceive coarse-grained entity information as well.
With interlaced self-attention layers in our PPO-TD, phrases are also enriched with object contexts by aggregating information from object tokens to further enhance their discriminative ability.
Finally, the inner product between phrase and image features is conducted to generate a binary mask for each phrase based on the pixel-phrase matching similarities.

In order to provide more precise object contexts for phrases, we also design a Phrase-Object Contrastive Loss (POCL) to refine the learning process of object tokens representations.
The Hungarian algorithm is first performed to obtain the bipartite matching between object tokens and ground-truth masks/classes.
Since the ground-truth masks/classes corresponding to each phrase are also available in the PNG annotations, we can find which object tokens are matched with each phrase.
Utilizing these matching relations as the supervision, we devise a contrastive loss on phrases and object tokens to increase the feature similarities between matched phrase-object pairs and decrease those between unmatched pairs.
Constrained by this loss, phrase features can aggregate object contexts from more relevant object tokens to enhance their discriminative ability.

Our contributions are summarized as follows:
\begin{itemize}
    \item We propose to enrich phrases with coupled pixel and object contexts by a Phrase-Pixel-Object Transformer Decoder (PPO-TD), where both fine-grained part details and coarse-grained entity clues are aggregated to phrase features for comprehensive visual-linguistic interaction.
    \item We also design a Phrase-Object Contrastive Loss (POCL) to increase feature similarities between matched phrase-object pairs and decrease those between unmatched pairs, thus providing more precise object contexts for phrases.
    \item Extensive experiments on the PNG benchmark demonstrate our method achieves new state-of-the-art performance with large margins.
\end{itemize}

\begin{figure*}[t]
  \begin{center}
     \includegraphics[width=\linewidth]{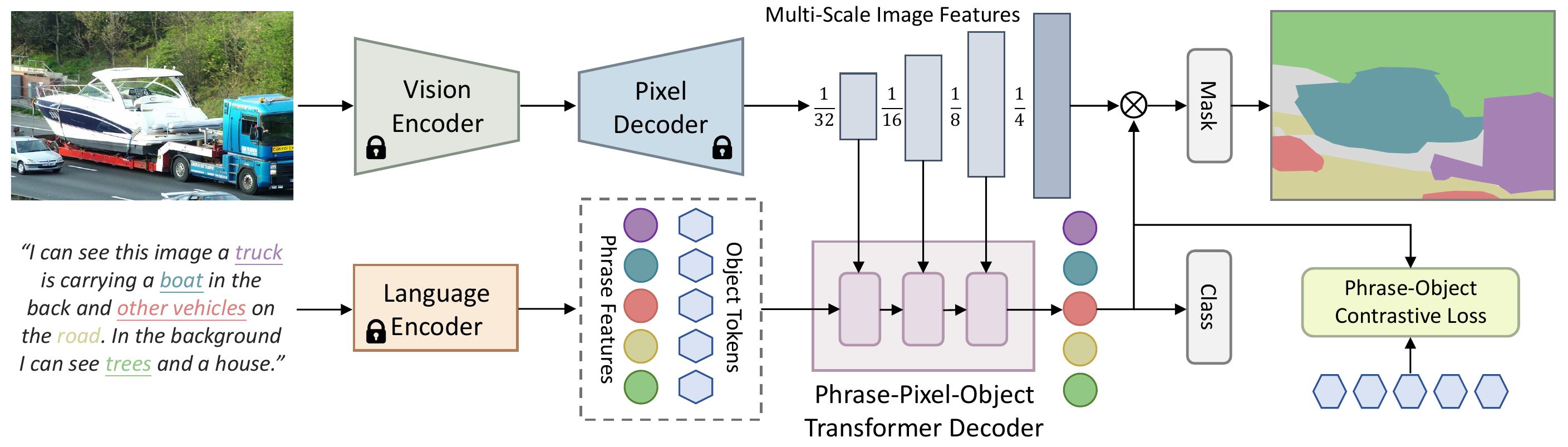}
  \end{center}
     \caption{Overview of our method. Image and phrase features are extracted by vision and language encoders. A pixel decoder further refines the multi-scale image features. Our proposed Phrase-Pixel-Object Transformer Decoder takes the concatenation of phrase features and learnable object tokens as queries and multi-scale image features as keys and values, where phrases are enriched with coupled pixel and object contexts to form a more comprehensive visual-linguistic interaction. A Phrase-Object Contrastive Loss is also proposed to increase feature similarities between matched phrase-object pairs and decrease those between unmatched ones so that more phrase-relevant object contexts are aggregated. The final mask predictions are obtained by the inner product between phrase and image features. The parameters of the language encoder, vision encoder, and pixel decoder are fixed during training.}
  \label{fig:framework}
\end{figure*}

\section{Related Work}

\subsection{Referring Image Segmentation}

The goal of referring image segmentation is to segment the object referred by the subject of an input sentence.
Previous mainstream methods~\cite{ye2019cross,huang2020referring,hui2020linguistic,liu2021cross,zhou2021attentive,tan2020multi,feng2021encoder,jing2021locate} are based on FCN~\cite{long2015fully} architecture and design various cross-modal attention mechanisms to fuse image and sentence features.
With the recent success of Transformer~\cite{vaswani2017attention}, LAVT~\cite{yang2022lavt} and ReSTR~\cite{kim2022restr} employ Vision Transformer~\cite{wang2022cris} to directly encode and fuse image and sentence features with additional channel attention or class seed embedding.
CRIS~\cite{wang2022cris} transfers large-scale vision-language pretrained knowledge of CLIP for text-to-pixel alignment.
PNG extends the grounding granularity of referring image segmentation from sentences to phrases, which could achieve fine-grained vision-language alignment.
In this paper, we propose to enrich phrases with coupled pixel and object contexts for more comprehensive visual-linguistic interaction.
The discussion of existing PNG methods has been included in Sec.~\ref{sec:intro}, so we omit it here.

\subsection{Panoptic Segmentation}

Panoptic segmentation task is first proposed by~\cite{kirillov2019ps}, whose aim is to unify semantic segmentation and instance segmentation tasks for a coherent scene segmentation of things and stuff classes.
Previous methods~\cite{kirillov2019pfpn,cheng2020panoptic,wang2020axial,mohan2021efficientps} mainly focus on designing specialized branches to tackle semantic and instance segmentation in a single network.
Recently, some Transformer-based methods~\cite{cheng2021per,zhang2021k,cheng2022masked} propose the mask classification paradigm to replace traditional per-pixel classification for more effective panoptic segmentation.
For example, MaskFormer~\cite{cheng2021per} exploits object queries to generate binary masks and their classifications simultaneously.
Different from panoptic segmentation, PNG segments objects according to narrative captions for a comprehensive cross-modal understanding.

\subsection{Phrase Grounding}

Phrase grounding requires the model to localize objects referred by noun phrases with bounding boxes.
Some early works~\cite{wang2018learning,plummer2018conditional,akbari2019multi} first extract features of region proposals and phrases separately, then perform phrase-region matching in common feature space with specially designed loss functions.
The following methods~\cite{bajaj2019g3raphground,mu2021disentangled,yu2020cross,dogan2019neural} focus on designing effective interaction approaches between visual and linguistic features to excavate multimodal context for more accurate phrase-region matching.
The recent GLIP model~\cite{li2022grounded} unifies object detection and phrase grounding into vision-language pretraining, which further boosts the grounding performance.
The PNG task also extends phrase grounding from region-level localization to pixel-level segmentation, which facilitates accurate object perception.

\section{Method}

The overall framework of our method is shown in Figure~\ref{fig:framework}, which follows the meta-architecture of Mask2Former~\cite{cheng2022masked}.
The input image and caption are first encoded by the vision and language encoders to obtain the image and phrase features.
Then, the pixel decoder extracts multi-scale image features after the vision encoder.
Our proposed Phrase-Pixel-Object Transformer Decoder (PPO-TD) takes the concatenation of phrase features and learnable object tokens as queries and multi-scale image features as keys and values.
Through successive layers in PPO-TD, phrases are enriched with coupled pixel and object contexts to form more comprehensive visual-linguistic interaction.
The inner product between phrase features and image features generates the final binary mask for each phrase.
Moreover, we also design a contrastive loss on phrase-object pairs to aggregate more precise object contexts for phrases.

\subsection{Multimodal Encoders and Pixel Decoder}

For the input image $\mathcal{I} \in \mathbb{R}^{H^0 \times W^0 \times 3}$, we adopt ResNet-50~\cite{he2016deep} as the vision encoder to extract image features $\bm{F}^i \in \mathbb{R}^{H^i \times W^i \times C^i_f}, i \in \{2,3,4,5\}$, where $H^i = \frac{H^0}{2^i}, W^i = \frac{W^0}{2^i}, C^i_f$ denote the height, width, and channel numbers of the $i$-th level of image features from the ResNet backbone.
For the input narrative caption with $N$ noun phrases, we adopt the BERT-base model~\cite{devlin2018bert} to extract phrase feature embeddings $\bm{R} \in \mathbb{R}^{N \times C_r}$, where $C_r$ denotes the channel number of phrase features.

Following~\cite{cheng2022masked}, we exploit multi-scale deformable attention Transformer (MSDeformAttn)~\cite{zhu2020deformable} as the pixel decoder to fuse and refine multi-scale image features.
The pixel decoder takes $\{\bm{F}^i\}_{i=3}^5$ as inputs and first projects the channel numbers of these image features into the same value $C_h$ with $1 \times 1$ convolution layers.
Then, the projected image features, their generated sinusoids positional embedding~\cite{vaswani2017attention} $\{\bm{P}^i\}_{i=3}^5 \in \mathbb{R}^{H^i \times W^i \times C_h}$, and learnable scale-level embedding $\bm{S} \in \mathbb{R}^{3 \times C_h}$ are fed into the pixel decoder for multi-scale feature refinement:
\begin{equation}
    \{\bm{\bar{F}}^i\}_{i=3}^5 = {\rm{MSDeformAttn}}(\{\bm{F}^i\}_{i=3}^5, \{\bm{P}^i\}_{i=3}^5, \bm{S}).
\end{equation}
Within the pixel decoder, each pixel on the image features samples a small set of positions from features of all scales using learned offsets and aggregates their features using learned attention weights.
$\{\bm{\bar{F}}^i\}_{i=3}^5$ denote image features after multiple deformable attention layers, which fuse rich multi-scale visual information.
The resolutions of $\{\bm{\bar{F}}^i\}_{i=3}^5$ are $1/8$, $1/16$, and $1/32$ of the original image resolution, respectively.
In addition, a bilinear interpolation layer is applied on $\bm{\bar{F}}^3$ to upsample its resolution by $2 \times$, and the upsampled feature is summed and fused with input backbone feature $\bm{F}^2$ using lateral convolution layers:
\begin{equation}
    \bm{\bar{F}}^2 = {\rm{Conv_{3 \times 3}}}({\rm{Conv_{1 \times 1}}}(\bm{F}^2) + {\rm{Upsample}}(\bm{\bar{F}}^3))
\end{equation}
The obtained feature $\bm{\bar{F}}^2$ has $1/4$ scale of the original image and serves as the per-pixel embedding to generate binary masks for phrases and object tokens.

\subsection{Phrase-Pixel-Object Transformer Decoder}

In order to enhance the discriminative ability of phrase features for distinguishing referred objects and predicting accurate segmentation masks, we propose a Phrase-Pixel-Object Transformer Decoder (PPO-TD) to enrich phrases with coupled pixel and object contexts for more comprehensive visual-linguistic interaction.

Each layer of our PPO-TD is mainly composed of cross-attention, self-attention, and feedforward network layers in a sequential manner.
Given the effectiveness and popularity of cross-attention, previous PNG methods also leverage it to conduct visual-linguistic interaction.
They utilize phrase features alone as the query and mask proposal features or image features as the key and value to aggregate proposal contexts only~\cite{gonzalez2021panoptic} or pixel contexts only~\cite{ding2022ppmn} to the phrases.
However, proposal contexts may contain fragmentary and noisy visual object information due to the inaccurate segments predicted by the off-the-shelf panoptic segmentation model.
Meanwhile, only pixel contexts also tend to incur semantic inconsistency between abstract noun phrases and concrete image pixels since lacking object-level visual contexts may yield sub-optimal object delineation.

Therefore, we argue that both pixel contexts and object contexts are beneficial to enhancing the discriminative ability of phrase features.
To introduce proper object-level visual contexts, we take the concatenation of phrase features and a set of learnable object tokens as the query of cross-attention instead of using phrase features alone.
For key and value, we adopt multi-scale image features from the pixel decoder.

Take the $l$-th layer of our PPO-TD as an example.
The $N$ phrase features $\bm{R}_{l-1} \in \mathbb{R}^{N \times C_h}$ and $M$ object tokens $\bm{O}_{l-1} \in \mathbb{R}^{M \times C_h}$ from the $(l-1)$-th layer are first summed with their learnable positional embeddings, and then concatenated and projected by a linear transformation to obtain the input query feature $\bm{Q}_l \in \mathbb{R}^{(N+M) \times C_h}$ of cross-attention layer:
\begin{equation}
    \bm{Q}_l = [\bm{R}_{l-1} + \bm{P}_r; \bm{O}_{l-1} + \bm{P}_o]\bm{W}_q,
\end{equation}
where $[;]$ is concatenation and $\bm{W}_q$ is linear transformation parameter.
The key and value features are similarly obtained:
\begin{equation}
    \bm{K}_l = (\bm{\bar{F}}_l + \bm{P}_l + \bm{S}_l)\bm{W}_k,~~\bm{V}_l = (\bm{\bar{F}}_l + \bm{P}_l + \bm{S}_l)\bm{W}_v,
\end{equation}
where $\bm{\bar{F}}_l \in \mathbb{R}^{H^lW^l \times C_h}$ is one of the flattened multi-scale image features $\{\bm{\bar{F}}^i\}_{i=3}^5$ from the pixel decoder, $\bm{P}_l$ and $\bm{S}_l$ are learnable positional embedding and scale-level embedding.
Following~\cite{cheng2022masked}, we exploit masked cross-attention to constrain the attention scope of phrase features and object tokens as follows:
\begin{equation}
    \bm{X}_l = {\rm{Softmax}}(\bm{Q}_l\bm{K}_l^{\rm{T}} + \bm{A}_{l-1})\bm{V}_l + \bm{X}_{l-1},
\end{equation}
where $\bm{X}_{l-1}, \bm{X}_l \in \mathbb{R}^{(N+M) \times C_h}$ are concatenated phrase features and object tokens in adjacent layers.
$\bm{A}_{l-1}$ is the attention mask and its value on position $(x, y)$ is:
\begin{equation}
    \bm{A}_{l-1}(x,y) = 
        \begin{cases}
            0, &\text{if}~\bm{M}_{l-1}(x,y) = 1 \\
            -\infty, &\text{otherwise},
        \end{cases}
\end{equation}
where $\bm{M}_{l-1} = {\rm{Resize}}(\bm{X}_{l-1}(\bm{\bar{F}}^2)^{\rm{T}}) \in \mathbb{R}^{(N+M) \times H^lW^l}$ is the predicted binary masks of phrases and object tokens after normalization and binarization at threshold $0.5$ in the $l-1$-th layer of PPO-TD.
The attention mask constrains the cross-attention scope to predicted foreground regions, which facilitates the aggregation of more phrase-relevant pixel contexts to the phrase features, and also refines the representations of object tokens by suppressing other distractors.

After the cross-attention layer, concatenated phrase features and object tokens are fed into the self-attention layer to communicate information with each other.
By this means, phrase features can also aggregate object contexts from object tokens whose representations are adaptively extracted from multi-scale image features, while object tokens incorporate language priors to distinguish referred objects better.
With coupled pixel and object contexts, the discriminative ability of phrase features is further enhanced to form a more comprehensive visual-linguistic interaction.

\subsection{Phrase-Object Contrastive Loss}

After $L$ layers of PPO-TD, we take the output phrase features $\bm{R}_L$ to generate final mask prediction $\bm{M}_L \in \mathbb{R}^{N \times H^0W^0}$ by inner product with the flattened per-pixel embedding $\bm{\bar{F}}^2$ followed by upsampling operation.
Since each phrase is also associated with a ground-truth class to which the referred object belongs, we also use a linear layer to generate final class prediction $\bm{C}_L \in \mathbb{R}^{N \times K}$ to leverage class priors contained in phrases, where $K$ is the number of classes.
Then, the mask classification loss of phrases $\mathcal{L}_{\rm{mask\text{-}cls}}^{\rm{phr}}$, which consists of a cross-entropy classification loss and a binary cross-entropy mask loss, is formulated as follows:
\begin{equation}
    \mathcal{L}_{\rm{mask\text{-}cls}}^{\rm{phr}} = \sum_{j=1}^N\left[ -{\rm{log}}p(c^*_j) + \mathcal{L}_{\rm{mask}}(\bm{m}_j, \bm{m}^*_j) \right],
\end{equation}
where $c^*_j$ is the ground-truth class of the $j$-th phrase, $m_j, m^*_j \in \mathbb{R}^{H^0 \times W^0}$ are the predicted and ground-truth masks of the $j$-th phrase respectively.
For learnable object tokens, we optimize a bipartite matching~\cite{cheng2022masked} to find the matched ground-truth mask/class for each object token.
Similar mask classification loss $\mathcal{L}_{\rm{mask\text{-}cls}}^{\rm{obj}}$ is also applied to object tokens, but the predicted masks of object tokens are not used as the final output results during inference.

Since phrases and object tokens are both matched with their own ground-truth masks/classes, we use these ground truths as links to obtain the actual matching relations $\bm{G}^* \in \mathbb{R}^{N \times M}$ between phrases and object tokens.
Inner-product between features of phrases and object tokens is also performed to generate the matching predictions $\bm{G} \in \mathbb{R}^{N \times M}$.
Based on $\bm{G}^*$ and $\bm{G}$, we design a Phrase-Object Contrastive Loss $\mathcal{L}_{\rm{poc}}$ (POCL) to increase the feature similarities between matched phrase-object pairs and decrease those between unmatched pairs, which is implemented with binary cross-entropy loss:
\begin{equation}
    \begin{split}
    \mathcal{L}_{\rm{bce}}(\bm{G}_{j,k}, \bm{G}^*_{j,k}) = &-\bm{G}^*_{j,k}{\rm{log}}(\bm{G}_{j,k}) \\
    &- (1 - \bm{G}^*_{j,k}){\rm{log}}(1 - \bm{G}_{j,k}),
    \end{split}
\end{equation}
\begin{equation}
    \label{eq:pocl}
    \mathcal{L}_{\rm{poc}}(\bm{G}, \bm{G}^*) = \frac{1}{NM}\sum_{j=1}^N\sum_{k=1}^M\mathcal{L}_{\rm{bce}}(\bm{G}_{j,k}, \bm{G}^*_{j,k}).
\end{equation}
Supervised by $\mathcal{L}_{\rm{poc}}$, more precise object contexts from more phrase-relevant object tokens can be aggregated to phrase features, which further enhances their discriminative ability.
Our POCL is inspired by previous works~\cite{he2020momentum,pang2021quasi,wu2022defense,li2022video} which leverage similar loss functions for supervised contrastive learning, but POCL contains intuitive designs tailored for the PNG task where richer contexts are naturally coupled.

Also noteworthy is that Eq.~\ref{eq:pocl} can be rewritten as follows:
\begin{equation}
    \begin{split}
        \mathcal{L}_{\rm{poc}}(\bm{G}, \bm{G}^*) &= \frac{1}{N}\sum_{j=1}^N\frac{1}{M}\sum_{k=1}^M\mathcal{L}_{\rm{bce}}(\bm{G}_{j,k}, \bm{G}^*_{j,k}) \\
        &= \frac{1}{N}\sum_{j=1}^N\mathcal{L}_{\rm{multi\text{-}cls}}(\bm{G}_{j,:}, \bm{G}^*_{j,:}) \\
        &= \frac{1}{M}\sum_{k=1}^M\mathcal{L}_{\rm{multi\text{-}cls}}(\bm{G}_{:,k}, \bm{G}^*_{:,k}).
    \end{split}
\end{equation}
$\mathcal{L}_{\rm{poc}}$ is equivalent to bi-directional multi-label classification between phrases and object tokens, which means each phrase can be classified into multiple object token labels (plural noun) and each object token can also be classified into multiple phrase labels (same object referred by different phrases).
By this means, positive phrase-object pairs obtain high classification probabilities, while negative ones do the opposite.
The final loss of our model is the combination of three losses:
\begin{equation}
    \mathcal{L} = \mathcal{L}_{\rm{mask\text{-}cls}}^{\rm{phr}} + \mathcal{L}_{\rm{mask\text{-}cls}}^{\rm{obj}} + \mathcal{L}_{\rm{poc}}.
\end{equation}

\section{Experiments}

\subsection{Dataset and Evaluation Metrics}
We conduct experiments on the Panoptic Narrative Grounding (PNG) benchmark~\cite{gonzalez2021panoptic}.
PNG is constructed by transferring caption annotations of Localized Narratives dataset~\cite{pont2020connecting} to the panoptic segmentation annotations of MSCOCO dataset~\cite{lin2014microsoft}.
For the linguistic domain, each narrative in PNG contains an average of $11.3$ noun phrases, of which 5.1 noun phrases require to be grounded.
The total number of noun phrases is $726,445$.
For the visual domain, a total of $741,697$ segments ($659,298$ are unique) in the panoptic segmentation annotations of MSCOCO are matched with narrative captions.

Following prior works~\cite{gonzalez2021panoptic,ding2022ppmn}, we adopt Average Recall as the metric to evaluate our method.
For each phrase, different Intersection over Union (IoU) thresholds between segmentation prediction and ground truth are considered to determine whether the prediction is a true positive or not.
A curve of recall at different IoU thresholds is obtained to calculate the Average Recall, which is the area under the curve.
For each plural noun phrase, all ground-truth instances are aggregated into a single segmentation to calculate IoU with similarly aggregated prediction.

\subsection{Implementation Details}

Our model is implemented with PyTorch.
The vision encoder and pixel decoder are pretrained on MSCOCO panoptic segmentation task for a fair comparison with prior works~\cite{gonzalez2021panoptic,ding2022ppmn}.
The projected channel number $C_h$ in the pixel decoder and our PPO-TD is $256$.
The input image is resized to $1024 \times 1024$.
The language encoder is implemented with the base version of BERT~\cite{devlin2018bert} whose embedding dimension $C_r$ is $768$.
The input caption has a maximum length of $230$ words, among which at most $N=30$ noun phrases are required to be grounded.
The number of learnable object tokens $M=100$.
Our PPO-TD has $L=9$ layers, and the mask classification losses for phrases and object tokens are applied before each layer as auxiliary losses.
AdamW is used as the optimizer with a learning rate of $1e^{-4}$.
We train our model with batch size 16 for $145K$ iterations on eight NVIDIA V100 GPUs.
The parameters of the language encoder, vision encoder, and pixel decoder are fixed during training.

\begin{figure*}[t]
   \centering
   \medskip
   \begin{subfigure}[t]{.33\linewidth}
  \centering
  \resizebox{0.95\linewidth}{!}{%
  \begin{tikzpicture}[/pgfplots/width=1.45\linewidth, /pgfplots/height=1.45\linewidth]
    \begin{axis}[
                 ymin=0,ymax=1,xmin=0,xmax=1,
        		 xlabel=IoU,
        		 ylabel=Recall@IoU,
         		 xlabel shift={-2pt},
        		 ylabel shift={-3pt},
		         font=\small,
		         axis equal image=true,
		         enlargelimits=false,
		         clip=true,
        	     grid style=solid, grid=both,
                 major grid style={white!85!black},
        		 minor grid style={white!95!black},
		 		 xtick={0,0.1,...,1.1},
                 xticklabels={0,.1,.2,.3,.4,.5,.6,.7,.8,.9,1},
        		 ytick={0,0.1,...,1.1},
                 yticklabels={0,.1,.2,.3,.4,.5,.6,.7,.8,.9,1},
         		 minor xtick={0,0.02,...,1},
		         minor ytick={0,0.02,...,1},
        		 legend style={at={(0.05,0.05)},
                 		       anchor=south west},
                 legend cell align={left}]
    \addplot+[green,solid,mark=none,ultra thick] table[x=IoU,y=PPO_overall]{figures/figure4_final.txt};
    \addlegendentry{Ours}
    \addplot+[red,solid,mark=none,ultra thick] table[x=IoU,y=ppm_overall]{figures/figure1_final.txt};
    \addlegendentry{PPMN}
    \addplot+[blue,solid,mark=none,ultra thick] table[x=IoU,y=PNG]{figures/figure1_final.txt};
    \addlegendentry{PNG}
    \end{axis}
\end{tikzpicture}}
  \subcaption{Overall performance}
  \label{fig:overall}
\end{subfigure}
\begin{subfigure}[t]{.33\linewidth}
  \centering
    \resizebox{0.95\linewidth}{!}{%
  \begin{tikzpicture}[/pgfplots/width=1.45\linewidth, /pgfplots/height=1.45\linewidth]
    \begin{axis}[
                 ymin=0,ymax=1,xmin=0,xmax=1,
        		 xlabel=IoU,
        		 ylabel=Recall@IoU,
         		 xlabel shift={-2pt},
        		 ylabel shift={-3pt},
		         font=\small,
		         axis equal image=true,
		         enlargelimits=false,
		         clip=true,
        	     grid style=solid, grid=both,
                 major grid style={white!85!black},
        		 minor grid style={white!95!black},
		 		 xtick={0,0.1,...,1.1},
                 xticklabels={0,.1,.2,.3,.4,.5,.6,.7,.8,.9,1},
        		 ytick={0,0.1,...,1.1},
                 yticklabels={0,.1,.2,.3,.4,.5,.6,.7,.8,.9,1},
         		 minor xtick={0,0.02,...,1},
		         minor ytick={0,0.02,...,1},
        		 legend style={at={(0.05,0.05)},
                 		       anchor=south west},
                 legend cell align={left}]
    \addplot+[green,dashed,mark=none,ultra thick] table[x=IoU,y=PPO_things]{figures/ppotd_new.txt};
    \addlegendentry{Ours (Things)}
    \addplot+[green,solid,mark=none,ultra thick] table[x=IoU,y=PPO_stuff]{figures/ppotd_new.txt};
    \addlegendentry{Ours (Stuff)}
    \addplot+[red,dashed,mark=none,ultra thick] table[x=IoU,y=ppm_things]{figures/figure2_final.txt};
    \addlegendentry{PPMN (Things)}
    \addplot+[red,solid,mark=none,ultra thick] table[x=IoU,y=ppm_stuff]{figures/figure2_final.txt};
    \addlegendentry{PPMN (Stuff)}
    \addplot+[olive,dashed,mark=none,ultra thick] table[x=IoU,y=MCN_things]{figures/figure2_final.txt};
    \addlegendentry{MCN (Things)}
    \addplot+[blue,dashed,mark=none,ultra thick] table[x=IoU,y=PNG_things]{figures/figure2_final.txt};
    \addlegendentry{PNG (Things)}
    \addplot+[blue,solid,mark=none,ultra thick] table[x=IoU,y=PNG_stuff]{figures/figure2_final.txt};
    \addlegendentry{PNG (Stuff)}
    \end{axis}
\end{tikzpicture}}
  \subcaption{Things and stuff categories}
  \label{fig:things_stuff}
\end{subfigure}
\begin{subfigure}[t]{.33\linewidth}
  \centering
      \resizebox{0.95\linewidth}{!}{%
  \begin{tikzpicture}[/pgfplots/width=1.45\linewidth, /pgfplots/height=1.45\linewidth]
    \begin{axis}[
                 ymin=0,ymax=1,xmin=0,xmax=1,
        		 xlabel=IoU,
        		 ylabel=Recall@IoU,
         		 xlabel shift={-2pt},
        		 ylabel shift={-3pt},
		         font=\small,
		         axis equal image=true,
		         enlargelimits=false,
		         clip=true,
        	     grid style=solid, grid=both,
                 major grid style={white!85!black},
        		 minor grid style={white!95!black},
		 		 xtick={0,0.1,...,1.1},
                 xticklabels={0,.1,.2,.3,.4,.5,.6,.7,.8,.9,1},
        		 ytick={0,0.1,...,1.1},
                 yticklabels={0,.1,.2,.3,.4,.5,.6,.7,.8,.9,1},
         		 minor xtick={0,0.02,...,1},
		         minor ytick={0,0.02,...,1},
        		 legend style={at={(0.05,0.05)},
                 		       anchor=south west},
                 legend cell align={left}]
    \addplot+[green,dashed,mark=none,ultra thick] table[x=IoU,y=PPO_singulars]{figures/ppotd_new.txt};
    \addlegendentry{Ours (Singulars)}
    \addplot+[green,solid,mark=none,ultra thick] table[x=IoU,y=PPO_plurals]{figures/ppotd_new.txt};
    \addlegendentry{Ours (Plurals)}
    \addplot+[red,dashed,mark=none,ultra thick] table[x=IoU,y=ppm_singulars]{figures/figure3_final.txt};
    \addlegendentry{PPMN (Singulars)}
    \addplot+[red,solid,mark=none,ultra thick] table[x=IoU,y=ppm_plurals]{figures/figure3_final.txt};
    \addlegendentry{PPMN (Plurals)}
    \addplot+[blue,dashed,mark=none,ultra thick] table[x=IoU,y=PNG_singulars]{figures/figure3_final.txt};
    \addlegendentry{PNG (Singulars)}
    \addplot+[blue,solid,mark=none,ultra thick] table[x=IoU,y=PNG_plurals]{figures/figure3_final.txt};
    \addlegendentry{PNG (Plurals)}
    \end{axis}
\end{tikzpicture}}
  \subcaption{Singulars and plurals}
  \label{fig:singulars_plurals}
\end{subfigure}
   \caption{\textbf{Average Recall Curves} for our method performance (a) compared to the state-of-the-art methods, disaggregated into (b) things and stuff categories, and (c) singulars and plurals noun phrases.}
  \label{figures:average_recall_curve}
 \end{figure*}
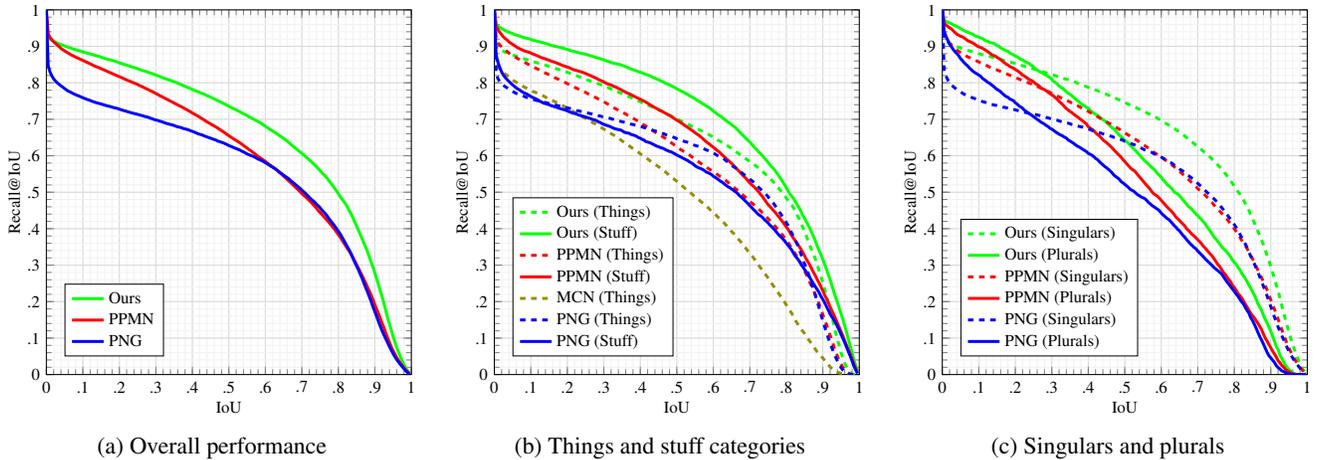

\begin{table*}[t]
\begin{subtable}{.5\linewidth}
  \centering
  \caption{Things and stuff categories.}
    \label{tab:things_stuff_results}
    {\tablestyle{5pt}{1}{
    \begin{tabular}{r||c|c|c}
    \hline\thickhline
    \rowcolor{mygray}
    & \multicolumn{3}{c}{Average Recall} \\
    \rowcolor{mygray}
    \multirow{-2}*{Method} & \multicolumn{1}{c}{overall} & \multicolumn{1}{c}{things} & \multicolumn{1}{c}{stuff} \\ \hline\hline
    PNG~\shortcite{gonzalez2021panoptic} & 55.4 & 56.2 & 54.3 \\
    MCN~\shortcite{luo2020multi} & - & 48.2 & - \\ 
    PPMN~\shortcite{ding2022ppmn} & 59.4 & 57.2  & 62.5 \\ 
    \hline\hline
    Ours & \textbf{66.1\color{sota_blue}{({+6.7})}} & \textbf{63.3\color{sota_blue}{({+6.1})}} & \textbf{69.8\color{sota_blue}{({+7.3})}} \\
    \hline
    \end{tabular}
    }}
\end{subtable}%
\begin{subtable}{.5\linewidth}
  \centering
  \caption{Singulars and plurals noun phrases.}
    \label{tab:singulars_plurals_results}
  {\tablestyle{5pt}{1}{
    \begin{tabular}{r||c|c|c}
    \hline\thickhline
    \rowcolor{mygray}
    & \multicolumn{3}{c}{Average Recall} \\
    \rowcolor{mygray}
    \multirow{-2}*{Method} & \multicolumn{1}{c}{overall} & \multicolumn{1}{c}{singulars} & \multicolumn{1}{c}{plurals} \\ \hline\hline
    PNG~\shortcite{gonzalez2021panoptic} & 55.4 & 56.2 & 48.8 \\ 
    PPMN~\shortcite{ding2022ppmn} & 59.4 & 60.0 & 54.0 \\ \hline\hline
    Ours & \textbf{66.1\color{sota_blue}{({+6.7})}} & \textbf{66.9\color{sota_blue}{({+6.9})}} & \textbf{58.6\color{sota_blue}{({+4.6})}} \\
    \hline
    \end{tabular}
    }}
\end{subtable}
\caption{Comparison with previous state-of-the-art methods on the PNG benchmark, disaggregated into (a) things and stuff categories, and (b) singulars and plurals noun phrases.}
\label{tab:sota_results}
\end{table*}

\subsection{Comparison with State-of-the-Art Methods}
We compare our method with previous state-of-the-art methods on the PNG benchmark.
Since PNG~\cite{gonzalez2021panoptic} is a pretty new task, only several previous methods could be compared.
Table~\ref{tab:sota_results} summarizes the quantitative comparison results.
It is observed that our method outperforms previous state-of-the-art ones by significant margins.
Compared with the latest one-stage method PPMN~\cite{ding2022ppmn}, our method achieves significant absolute performance gains of $6.7/6.1/7.3/6.9/4.6$ on Average Recall metric for \textit{overall/things/stuff/singulars/plurals} splits, which demonstrates enriching phrases with coupled pixel and object contexts is superior to aggregating pixel contexts only due to the coarse-grained entity clues complemented by object tokens.
Compared with the two-stage baseline method proposed by PNG, our performance boost becomes $10.7$ on Average Recall for \textit{overall} split, which shows the offline-generated mask proposals may introduce noisy and fragmentary object-level contexts while our learned object tokens can provide more precise object contexts to phrases.
MCN~\cite{luo2020multi} tackles referring expression segmentation and comprehension with a collaborative network and its Average Recall is evaluated on only \textit{things} split.
Both our method and previous PNG methods largely outperform MCN, which is probably because sentence-mask/box alignment in referring tasks is insufficient to learn phrase-level segmentation well.

As illustrated in Figure~\ref{figures:average_recall_curve}, we also draw the recall curves of our methods and previous ones for detailed comparison.
When the IoU thresholds are around $0$, the recall values of all methods approach $1$.
However, the curves of our method are consistently above those of previous counterparts at most IoU thresholds, showing that our method consistently grounds more noun phrases under different evaluation conditions.
In particular, the areas under our curves are still larger than those of other methods when the IoU threshold is beyond $0.9$, which indicates our method can produce more accurate segmentation masks upon very strict metrics.

\begin{figure*}[t]
  \begin{center}
     \includegraphics[width=0.9\linewidth]{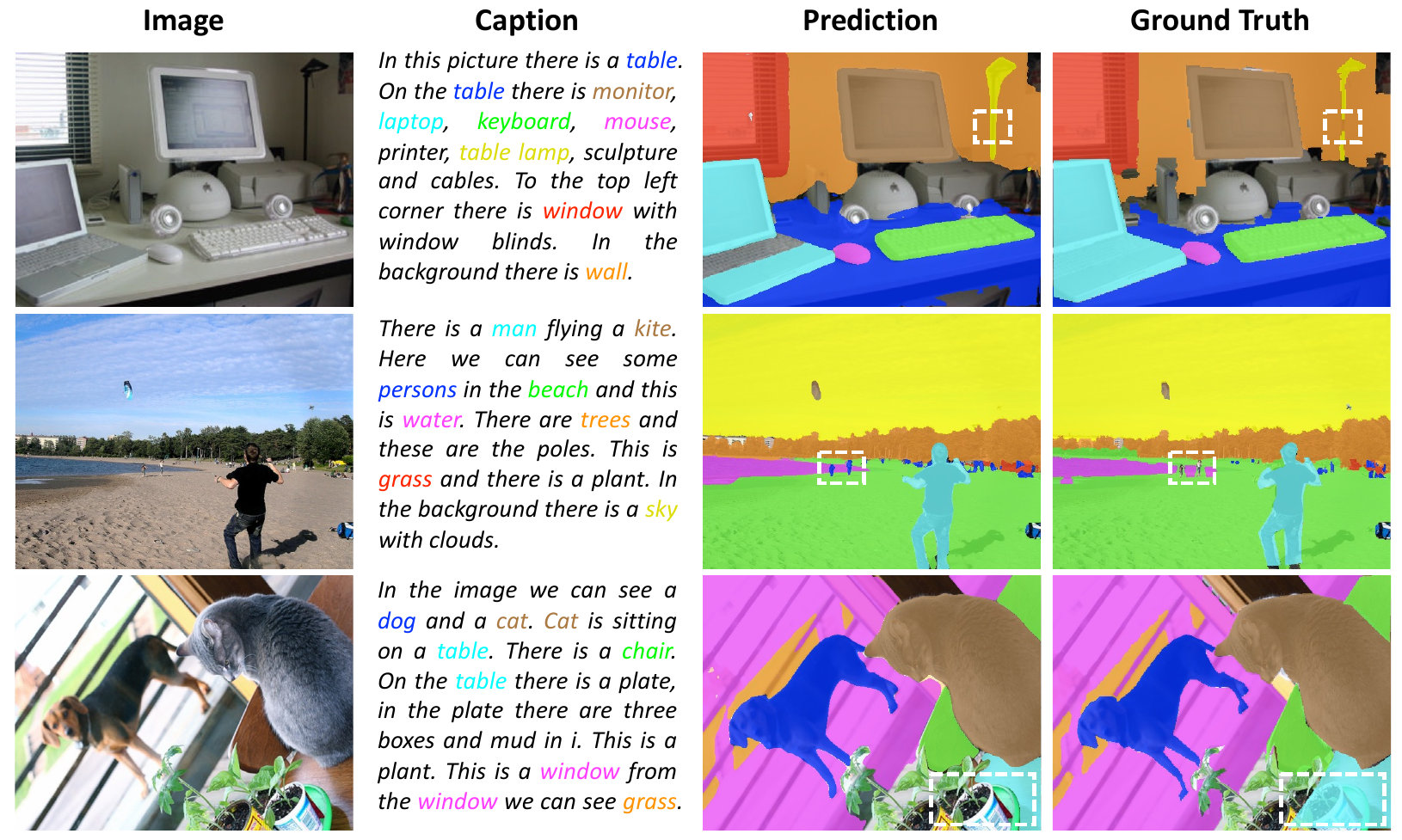}
  \end{center}
     \caption{Qualitative results of our method. The colors of phrases correspond to those of segments.}
  \label{fig:qualitative}
\end{figure*}

\begin{table}[!htbp]
    \centering
    {
    \tablestyle{3.3pt}{1}\begin{tabular}{c|c|c|c||c|c|c|c|c}
    \hline\thickhline
    \rowcolor{mygray}
    \multicolumn{3}{c|}{PPO-TD} & & \multicolumn{5}{c}{Average Recall} \\
    \rowcolor{mygray}
    \multicolumn{1}{c}{PC} & \multicolumn{1}{c}{MA} & \multicolumn{1}{c|}{OC} & \multirow{-2}*{POCL} &\multicolumn{1}{c}{overall} & \multicolumn{1}{c}{things} & \multicolumn{1}{c}{stuff} & \multicolumn{1}{c}{singulars} & \multicolumn{1}{c}{plurals} \\ \hline\hline
    \checkmark & & & & 64.7 & 61.6 & 68.8 & 65.6 & 56.1 \\
    \checkmark & \checkmark & & & 64.9 & 61.7 & 69.1 & 65.8 & 56.4 \\
    \checkmark & \checkmark & \checkmark & & 65.5 & 62.6 & 69.3 & 66.4 & 57.0 \\
    \checkmark & \checkmark & \checkmark & \checkmark & \textbf{66.1} & \textbf{63.3} & \textbf{69.8} & \textbf{66.9} & \textbf{58.6} \\
    \hline
    \end{tabular}
    }
    \caption{Verifying the effectiveness of components in our proposed PPO-TD and POCL. ``PC'': pixel contexts. ``MA'': masked attention. ``OC'': object contexts.}
    \label{tab:ablation:componenets}
\end{table}

\subsection{Ablation Studies}

We also conduct ablation studies on the PNG benchmark to verify the effectiveness of our network designs.

\textbf{Component Analysis.}
In Table~\ref{tab:ablation:componenets}, we analyze the effects of our proposed components.
The $1$-st row denotes aggregating pixel contexts (PC) to phrases by standard cross-attention in our proposed PPO-TD, which has achieved considerably higher performance than previous methods.
The result of PC shows that pixel contexts are necessary to PNG models for generating masks of high quality.
The $2$-nd row represents applying masked cross-attention (MA) between phrases and image features to aggregate pixel contexts from foreground regions.
The improvement indicates that MA can suppress background noises to provide more phrase-relevant pixel contexts.
In the $3$-rd and $4$-th row, OC denotes introducing learnable object tokens to provide object contexts for phrases, and POCL is the phrase-object contrastive loss which makes phrases obtain more precise object contexts.
Combining OC and POCL can achieve a further performance boost of $1.2$ Average Recall for \textit{overall} split over the particularly strong model in the $2$-nd row, which is very challenging.
This result well demonstrates that enriching phrases with coupled pixel and object contexts can further improve the strong performance of pixel contexts only, which shows the effectiveness of introducing object contexts.

\begin{table}[!htbp]
    \centering
    {
    \tablestyle{6pt}{1}\begin{tabular}{c||c|c|c|c|c}
    \hline\thickhline
    \rowcolor{mygray}
    & \multicolumn{5}{c}{Average Recall} \\
    \rowcolor{mygray}
    \multirow{-2}*{$L$} & \multicolumn{1}{c}{overall} & \multicolumn{1}{c}{things} & \multicolumn{1}{c}{stuff} & \multicolumn{1}{c}{sigulars} & \multicolumn{1}{c}{plurals} \\ \hline\hline
    3 & 65.0 & 61.8 & 69.5 & 66.0 & 56.5 \\
    6 & 65.5 & 62.5 & 69.5 & 66.4 & 56.9 \\
    9 & \textbf{66.1} & \textbf{63.3} & \textbf{69.8} & \textbf{66.9} & \textbf{58.6} \\
    \hline
    \end{tabular}
    }
    \caption{Results of different numbers of PPO-TD layers $L$.}
    \label{tab:ablation:layers-number}
\end{table}

\textbf{Number of PPO-TD layers.}
Table~\ref{tab:ablation:layers-number} shows the results of our PPO-TD with different numbers of layers.
Since three scales of image features are grouped into a stage in PPO-TD for three layers, we evaluate the choices of $3$, $6$, and $9$ layers.
we can find that more layers in PPO-TD yield higher performance, which shows that iteratively enriching phrases with pixel and object contexts is beneficial.

\begin{table}[!htbp]
    \centering
    {
    \tablestyle{6pt}{1}\begin{tabular}{c||c|c|c|c|c}
    \hline\thickhline
    \rowcolor{mygray}
    & \multicolumn{5}{c}{Average Recall} \\
    \rowcolor{mygray}
    \multirow{-2}*{Lang. Enc.} & \multicolumn{1}{c}{overall} & \multicolumn{1}{c}{things} & \multicolumn{1}{c}{stuff} & \multicolumn{1}{c}{sigulars} & \multicolumn{1}{c}{plurals} \\ \hline\hline
    BERT~\shortcite{devlin2018bert} & 66.1 & 63.3 & \textbf{69.8} & 66.9 & \textbf{58.6} \\
    CLIP~\shortcite{radford2021learning} & \textbf{66.2} & \textbf{63.7} & 69.5 & \textbf{67.0} & 58.4 \\
    T5~\shortcite{raffel2020exploring} & \textbf{66.2} & 63.6 & 69.6 & \textbf{67.0} & 58.5 \\
    \hline
    \end{tabular}
    }
    \caption{Results of different language encoders whose parameters are all fixed during training.}
    \label{tab:ablation:text-encoder}
\end{table}

\textbf{Different Language Encoders.}
Since our language encoder is fixed during training, we also explore the effects of using different language encoders to extract the phrase features.
As shown in Table~\ref{tab:ablation:text-encoder}, both the large-scale vision-language pretraining model CLIP~\cite{radford2021learning} and large-scale language pretraining model T5~\cite{raffel2020exploring} yield similar performance to BERT, which indicates our method is insensitive to the structure of language encoder if its parameters are fixed during training.

\subsection{Qualitative Results}
We present the qualitative results of our method in Figure~\ref{fig:qualitative}.
Our method can generate high-quality segmentation according to dense phrases.
We use white dashed boxes to mark some regions where our predictions are inconsistent with the ground truths.
For example, in the $2$-nd row, our method segments two people near the water while the ground-truth annotation is missing.
In the $3$-rd row, the ground truth annotates part of the plant pot as ``table'' while our method yields the correct prediction on the actual table area.

\section{Conclusion}

In this paper, we focus on the panoptic narrative grounding (PNG) task whose goal is to segment objects according to noun phrases in a narrative caption.
An essential factor of PNG is visual-linguistic interaction, where previous methods aggregate proposal contexts or pixel contexts to phrases for realizing it.
However, fragmentary visual information or semantic misalignment may be introduced by these methods and their performances are hence limited.
Therefore, we propose to enrich phrases with coupled pixel and object contexts with a Phrase-Pixel-Object Transformer Decoder (PPO-TD) for a more comprehensive visual-linguistic interaction.
In addition, we also propose a Phrase-Object Contrastive Loss (POCL) to aggregate more precise object contexts from more phrase-relevant object tokens.
Extensive experiments on the PNG benchmark show that our method outperforms previous state-of-the-art methods by large margins.

\appendix

\section*{Acknowledgments}

This research was supported in part by National Key R\&D Program of China (2022ZD0115502), National Natural Science Foundation of China (Grant No. 62122010), Zhejiang Provincial Natural Science Foundation of China under Grant No. LDT23F02022F02, and Key Research and Development Program of Zhejiang Province under Grant 2022C01082.

\bibliographystyle{named}
\bibliography{ijcai23}

\end{document}